\documentclass[letterpaper, 10 pt, journal, twoside]{IEEEtran}

\ifCLASSINFOpdf
\else
\fi

\usepackage{graphicx}
\usepackage{caption}
\usepackage{amsmath,amssymb,enumerate}

\usepackage{setspace}
\usepackage{booktabs}
\usepackage[usenames,dvipsnames,svgnames,table]{xcolor}
\usepackage{mathtools}
\usepackage{algorithm, algorithmicx, algpseudocode}
\usepackage{blindtext}
\usepackage{mathrsfs}
\usepackage[mathscr]{euscript}
\usepackage{times}
\usepackage{cite} % to group references
\usepackage{multicol}
\definecolor{darkgreen}{rgb}{0,0.6,0}
\usepackage[bookmarks=true,colorlinks=true,pdfpagemode=UseNone,citecolor=black,linkcolor=black,urlcolor=BrickRed]{hyperref}
\usepackage[caption=false,font=footnotesize]{subfig}
\usepackage{amsfonts}
\usepackage[utf8]{inputenc}
\usepackage[T1]{fontenc}
\usepackage{textcomp}
\usepackage{arydshln}
\usepackage{balance}
\usepackage{soul}

\newtheorem{problem}{Problem}

\renewcommand{\thefigure}{\arabic{figure}}

\newcommand{\diag}{\mathop{\mathrm{diag}}}

\newcommand{\transpose}{\mathsf{T}}

\makeatletter
\newcommand{\mathleft}{\@fleqntrue\@mathmargin0pt}
\newcommand{\mathcenter}{\@fleqnfalse}
\makeatother

\newcommand{\cc}[1]{\textcolor{black}{#1}}

\begin{document}

\title{Convex Geometric Trajectory Tracking using Lie Algebraic MPC for Autonomous Marine Vehicles}

\author{Junwoo Jang, Sangli Teng, and Maani Ghaffari,~\IEEEmembership{IEEE Member}
        % John~Doe,~\IEEEmembership{Fellow}
        % and~Jane~Doe,~\IEEEmembership{Life~Fellow}% <-this % stops a space
    \thanks{Manuscript received: May, 15, 2023; Revised August 20, 2023; Accepted October, 2, 2023. This paper was recommended for publication by Editor Jaydev P. Desai upon evaluation of the Associate Editor and Reviewers’ comments. The work of Maani Ghaffari was supported by NSF under Award No. 2118818. \textit{(Corresponding author: Junwoo Jang})}
    \thanks{The authors are with the University of Michigan, Ann Arbor 48109, USA.
\\    {\tt\footnotesize (junwoo@umich.edu; sanglit@umich.edu; maanigj@umich. edu)}
    }% <-this % stops a space
    \thanks{Digital Object Identifier (DOI): see top of this page.}
}

\markboth{IEEE Robotics and Automation Letters. Preprint Version. Accepted October, 2023}
{Jang \MakeLowercase{\textit{et al.}}: Convex Geometric Trajectory Tracking for Autonomous Marine Vehicles} 
% \markboth{Journal of \LaTeX\ Class Files,~Vol.~18, No.~9, September~2020}%
% {How to Use the IEEEtran \LaTeX \ Templates}

\maketitle

\begin{abstract}
Controlling marine vehicles in challenging environments is a complex task due to the presence of nonlinear hydrodynamics and uncertain external disturbances. Despite nonlinear model predictive control (MPC) showing potential in addressing these issues, its practical implementation is often constrained by computational limitations. 
In this paper, we propose an efficient controller for trajectory tracking of marine vehicles by employing a convex error-state MPC on the Lie group. By leveraging the inherent geometric properties of the Lie group, we can construct globally valid error dynamics and formulate a quadratic programming-based optimization problem. Our proposed MPC demonstrates effectiveness in trajectory tracking through extensive-numerical simulations, including scenarios involving ocean currents. Notably, our method substantially reduces computation time compared to nonlinear MPC, making it well-suited for real-time control applications with long prediction horizons or involving small marine vehicles.

\end{abstract}

% Note that keywords are not normally used for peerreview papers.

\IEEEpeerreviewmaketitle

\section{Introduction}
\IEEEPARstart{M}{arine} vehicles have become increasingly important due to their diverse applications, such as underwater exploration \cite{sahoo2019advancements}, the oil and gas industry \cite{zagatti2018flatfish}, transportation and environmental monitoring \cite{wang2019roboat}. Advancements in automation technology have led to the development of more advanced marine vehicles capable of performing complex tasks in harsh and challenging environments. However, controlling these vehicles is still arduous due to their highly nonlinear dynamics from complex hydrodynamic interactions and uncertain external disturbances. Additionally, characteristics such as low controllability, low motion frequency, and long control signal response time can lead to unstable behavior or overshoot, posing potential risks in situations where precise control is crucial, like collision avoidance, station keeping, and docking \cite{sarda2016station}. 

While classical control methods like proportional-integral-derivative (PID) controllers have been widely used for controlling marine vehicles \cite{liu2016unmanned}, they struggle to operate effectively in narrow waterways or heavy traffic circumstances \cite{cho2020efficient}. As a result, modern control techniques such as model predictive control (MPC) have gained popularity in recent years \cite{shi2021advanced, wei2022mpc}. MPC is a powerful control approach that is capable of handling constraints, nonlinear dynamics, and disturbances. However, nonlinear MPC for marine vehicles requires solving complex optimization problems, which can be computationally demanding and difficult to implement in real-time applications.

With recent advancements in computational capabilities, direct nonlinear optimization using nonlinear MPC (NMPC) has been employed in real marine vehicle experiments \cite{wang2018design, guerreiro2014trajectory, liang2020nonlinear, wang2021adaptive}. However, real-time NMPC requires certain approximations. For instance, \cite{wang2018design} employs simplified hydrodynamics with first-order and diagonal damping force terms, and \cite{guerreiro2014trajectory, liang2020nonlinear, wang2021adaptive} limits the vehicle maneuverability by lowering control frequency and speed of the vehicle. Moreover, these studies focus only on 3D motion, which limits their applicability for general marine vehicles control, such as station-keeping with heave motion or underwater vehicle control. In the case of small marine vehicles that cannot accommodate high-end computing devices, there is a need to reduce computational demands significantly.

%% figure
\begin{figure}[t]
\centering
\includegraphics[scale=0.21]{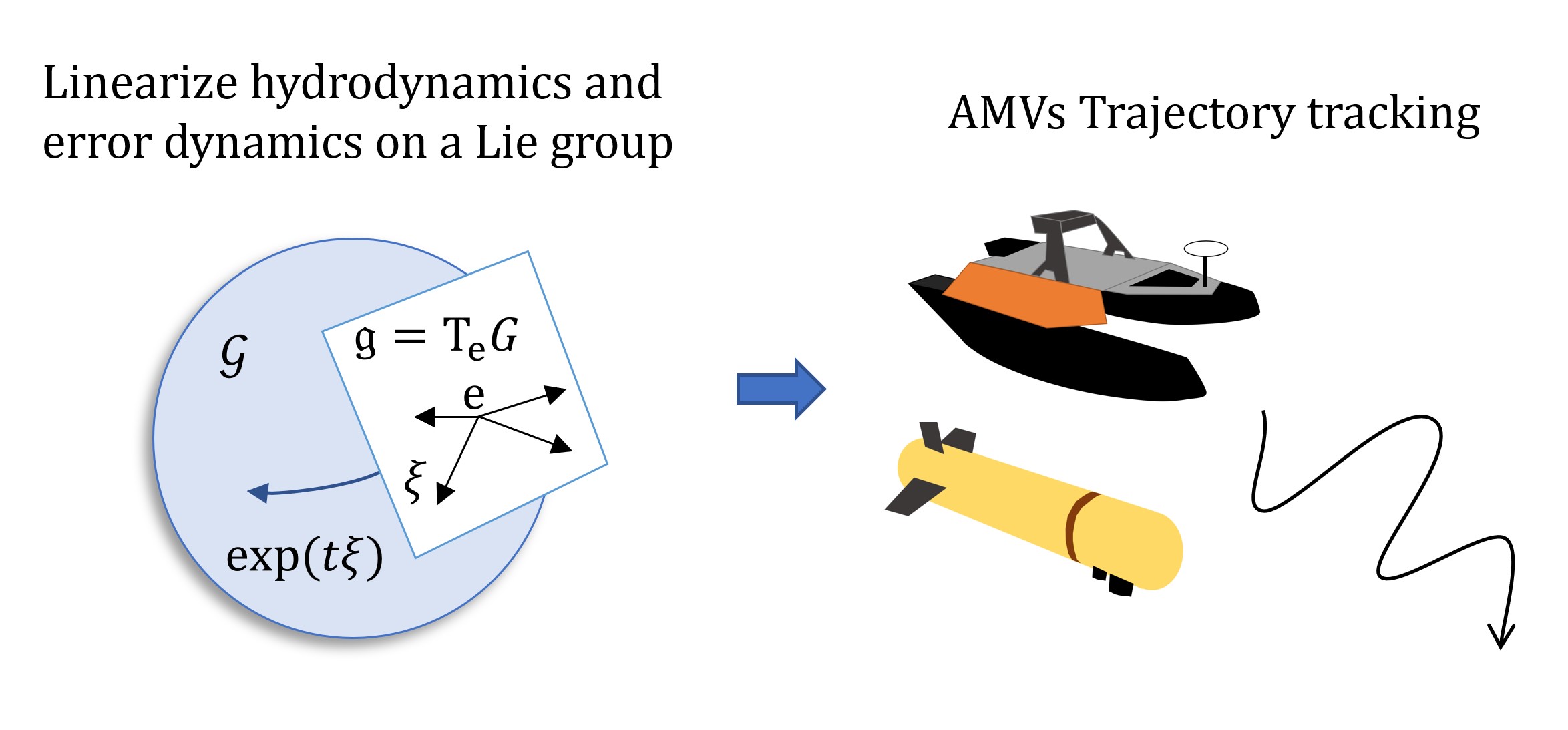}
\caption{\label{fig:framework} The proposed geometric trajectory tracking algorithm framework. The algorithm incorporates tracking error and hydrodynamics, which are defined on a Lie group and linearized to construct a convex MPC algorithm. The proposed MPC is applied to a marine vehicle within a simulation environment.}
% \vspace{-13pt}
\end{figure}

Computationally-efficient control algorithms capable of handling highly nonlinear hydrodynamics are crucial for marine vehicle control. Several promising approaches involve improved system representation using reasonable approximation or efficient control optimization to address this. Adaptive MPC \cite{annamalai2015robust} utilizes multiple approximated linear models to reduce the computational burden. MPC with projection neural network \cite{liu2020computationally} efficiently solves constrained optimization problems with parallel computational capability. Distributed optimization \cite{shen2020distributed} decomposes the original optimization problems into small subproblems by leveraging the dynamic properties of marine vehicle motion and then solving them with a significant reduction in computational complexity.

Another promising approach to achieving computational efficiency is geometric control, based on the Lie group framework, to exploit existing symmetry in the problem~\cite{teng2022lie,ghaffari2022progress}. Unlike methods that rely on approximations of the hydrodynamics model, geometric control leverages the intrinsic geometry of the system and represents the dynamics in an invariant and symmetric manner. Considering that the configuration of the vehicle space is a nonlinear manifold rather than a linear space, trajectory tracking algorithms for mobile robots and surface vehicles on $\mathrm{SE}(2)$ are presented in \cite{he2021point, tayefi2019logarithmic}. A recent study introduces error-state MPC on $\mathrm{SE}(3)$ for controlling legged robots \cite{teng2022error}, providing an accurate estimation of error dynamics. Geometric control guarantees that the error dynamics are globally valid and evolve independently of the system trajectory, enabling efficient quadratic programming (QP)-based control optimization. 

Motivated by the work of \cite{teng2022error}, we develop an error-state MPC on the Lie group for marine vehicle control, as illustrated in Fig. \ref{fig:framework}, constructing an efficient and accurate trajectory tracking controller. The marine domain imposes more challenging (and perhaps interesting) scenarios as the higher water density leads to significant environmental forces and state-dependent vehicle models. Our key contributions are summarized as follows.
\begin{enumerate}
\item We establish a nonlinear hydrodynamics model on the Lie group to ensure that error dynamics are globally valid and evolve independently of the system.
\item We develop a convex error-state MPC by employing first-order approximations of dynamics and error dynamics on the Lie group.
\item We demonstrate the effectiveness of the proposed algorithm in controlling surface vehicles for trajectory tracking in the presence of external disturbances using the Marine Systems Simulator \cite{fossen2004}. \cc{Our code is publicly available at
} \url{https://github.com/UMich-CURLY/Lie-MPC-AMVs}.
\end{enumerate}

The remainder of this paper is organized as follows. The dynamics of marine vehicles and its expression on the Lie group are presented in Section \ref{sec:Dynamics of Marine Vehicles}. The convex error-state MPC is derived from the linearization of dynamics and error dynamics on the Lie group in Section \ref{Geometric convex error-state MPC}. In Section \ref{sec:Numerical Simulations}, we present numerical simulations to evaluate the performance of our method in scenarios involving ocean currents and discuss potential directions for future research. Finally, we conclude the paper in Section \ref{sec:conclusion}. 

% \mgj{it's good to label the section and use Sec.~\ref{sec:conclusion} instead manually hardcoding. This becomes a problem for longer documents. I did the conclusion as an example.}

\section{Dynamics of Marine Vehicles}
\label{sec:Dynamics of Marine Vehicles}
% \mgj{It's better not to start with a subsection directly. You can briefly describe the purpose of the section and what is ahead (why and what or how).}
In this section, we present a background on the general hydrodynamics model of marine vehicles and Lie groups. Subsequently, we define the vehicle model within the framework of Lie groups.

\subsection{\cc{Marine vehicle equations of motion}}

Extensive research has been conducted to comprehend the hydrodynamics of marine vehicles and approximate the main forces acting on a vehicle. We follow Fossen's analytical approach \cite{fossen2011handbook} to select the most dominant forces and  model a marine vehicle.

In general, surface vehicles are modeled with 3\cc{-}degrees of freedom (DOF). However, for modeling and controlling a general marine vehicle, such as an autonomous underwater vehicle (AUV), we describe 6-DOF motion equations.

\begin{figure}[t]
\centering
\includegraphics[scale=0.18]{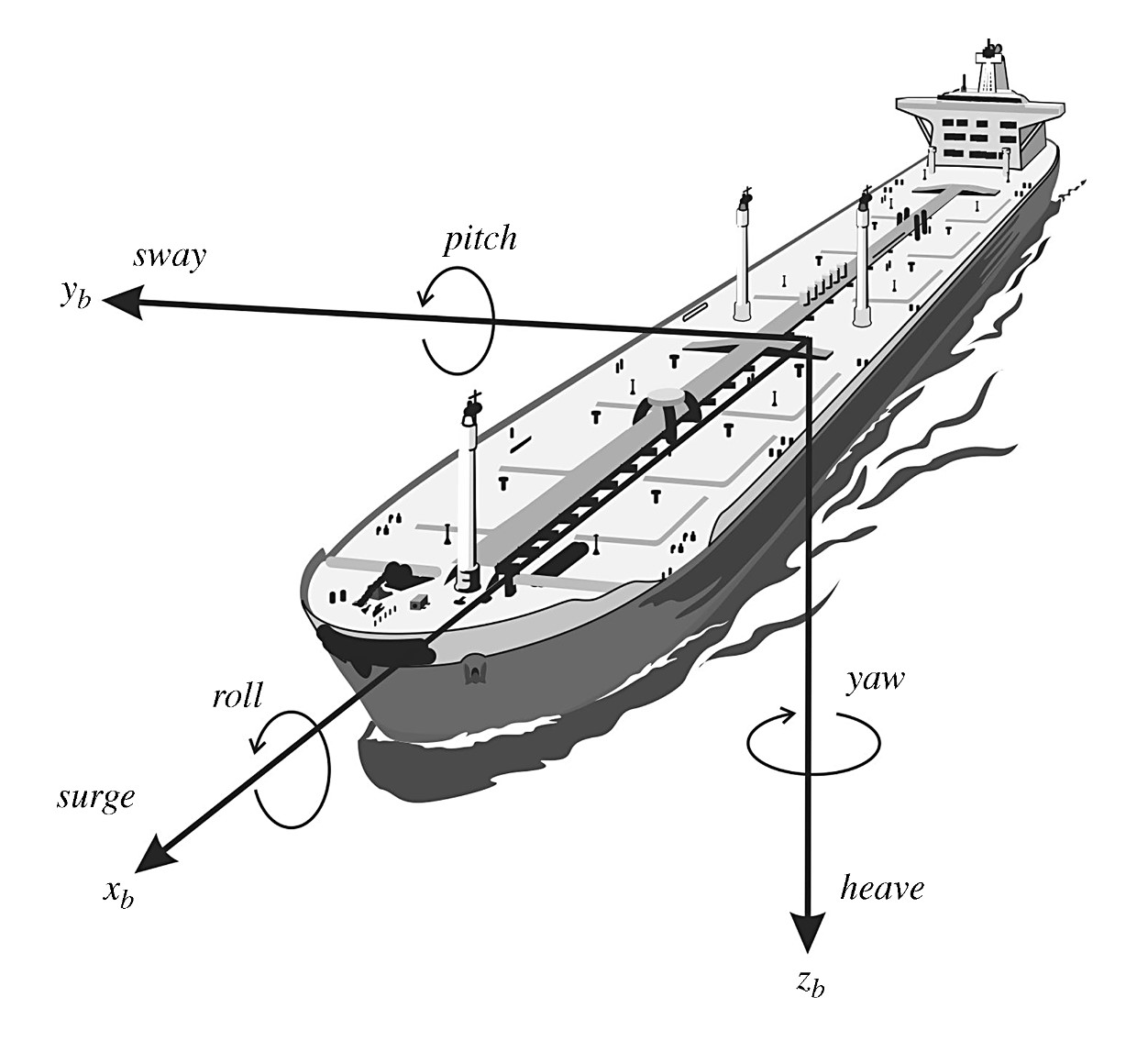}
\includegraphics[scale=0.18]{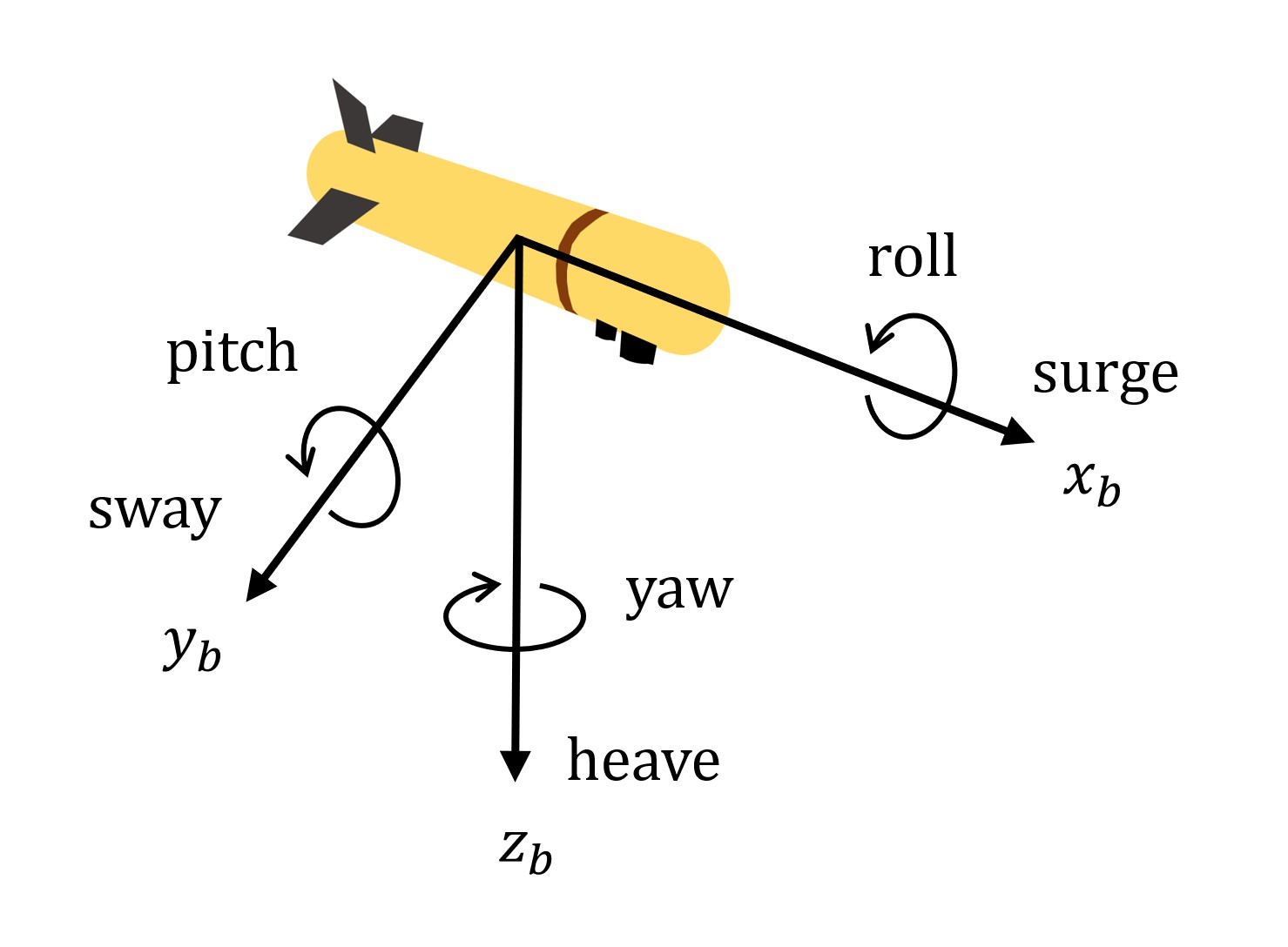}
\caption{\label{fig:axis} The 6\cc{-}DOF velocities in the body-fixed reference frame following Fossen's convention~\cite{fossen2011handbook}. 
% \mgj{Refer to all figures in the text near when they appear.}
}
% \vspace{-13pt}
\end{figure}

Let the rotation matrix $R \in \mathrm{SO}(3)$ and identity matrix $I_3 \in \mathbb{R}^{3 \times 3}$, where 
\begin{equation}
    \mathrm{SO}(3) = \{R| R \in \mathbb{R}^{3\times3}, RR^\transpose = R^\transpose R = I_3, \det(R) = 1\}.
\end{equation}
We use the notation $(\cdot)^\wedge$ to represent the cross-product operation, where $\lambda \times a := \lambda^\wedge a$. $\lambda^\wedge$ is a skew-symmetric matrix,
\begin{equation}
        \lambda^\wedge = \begin{bmatrix}
0 & -\lambda_z   & \lambda_y  \\
\lambda_z & 0   & -\lambda_x \\
-\lambda_y & \lambda_x   & 0,
\end{bmatrix} .
\end{equation}
The transformation matrix for converting from body-fixed frame to spatial frames is given by:
\begin{equation}
\dot{\eta} = J_{\Theta}(\nu) \nu ,
\end{equation}
where $J_{\Theta}$ is the Euler angle transformation matrix for 6-DOF kinematic equations, the vector $\nu$ includes body-frame velocity and angular velocity, $\nu = [u,v,w,p,q,r]^\transpose$ and the vector $\eta$ is the generalized position and orientation in North-East-Down (NED) frame, $\eta = [x, y, z, \phi, \theta, \psi]^\transpose$. For a marine vehicle, the six different motion components are conveniently defined as shown in Fig. \ref{fig:axis}.

In Fossen's model, the equations of motion for a marine vehicle in the body frame are expressed as a set of equations in the following form:
\begin{align}
\begin{split}
M_{RB}\dot{\nu} + C_{RB}(\nu)\nu + M_{AM}\dot{\nu}_r + C_{AM}(\nu_r) \nu_r  \\  + D(\nu_r) \nu_r + g(\eta) = \tau_c + \tau_{wind}  + \tau_{waves} ,
\end{split}
\end{align}
where $M$ is the mass matrix, $C(v)$ is the Coriolis matrix, $D(v)$ is the drag (damping) matrix, and $\tau$ is the vector of forces and moments. The subscripts $RB$ and $AM$ associated with the mass and Coriolis matrices denote the rigid body and additional mass, respectively. The added mass is assumed to be proportional only to the relative velocity $\nu_r$, which is commonly used for marine vehicle models. The force $\tau_c$ is generated by the propulsion system, which is the control input, and $\tau_{wind}$ and $\tau_{wave}$ are the external disturbances caused by the wind and waves. The vector $\nu_r$ is the relative velocity vector to the ocean current velocity $\nu_c$ (i.e., $\nu_r = \nu - \nu_c$). Note that terms relating to $\nu_r$ represent hydrodynamic forces, and $g(\eta)$ is the hydrostatic force.

The rigid body mass matrix is symmetric and represented as:

\begin{align}
\begin{split}
    M_{RB} &= \begin{bmatrix}
        m & 0 & 0 & 0 & mz_g & -my_g \\
        0 & m & 0 & -mz_g & 0 & mx_g \\
        0 & 0 & m & my_g & -mx_g & 0 \\
        0 & -mz_g & my_g & I_xx & -I_{xy} & -I_{xz} \\
        mz_g & 0 & -mx_g & -I_{yx} & I_{yy} & -I_{yz} \\
        -my_g & mx_g & 0 & -I_{zx} & -I_{zy} & I_{zz} \\
    \end{bmatrix}\\
    &= \begin{bmatrix}
        M_{11} & M_{12} \\
        M_{21} & M_{22}
    \end{bmatrix},
    \end{split}
\end{align}
where $x_g, y_g$, and $z_g$ are the distances to the center of gravity from the origin coordinate, and $I$ is the inertia dyadic about the origin coordinate. In an ideal fluid, the hydrodynamic system's added inertia matrix at infinite frequency is represented as positive definite and constant.

The Coriolis matrix captures the inertial forces in the dynamics of marine vehicles and is skew-symmetric. The Coriolis matrix is derived from the mass matrix and the relative velocity vector,
\begin{equation}
\begin{split}
    &C(\nu) = \\ 
    &\;\;\;\;\;\;\;\;\; \begin{bmatrix}
        0 & -(M_{11}\nu_1 + M_{12} \nu_2)^\wedge \\
        -(M_{11}\nu_1 + M_{12} \nu_2)^\wedge & -(M_{21}\nu_1 + M_{22} \nu_2)^\wedge
    \end{bmatrix} ,
    \end{split}
    \label{eq:coriolis}
\end{equation}
where $\nu_1 = [u,v,w]^\transpose$ and $\nu_2 = [p,q,r]^\transpose$.

% The motion of a marine vehicle moving in 6-DOF at high speed is highly nonlinear and coupled. 

% The Coriolis matrix of the added mass matrix can be equivalently calculated using (\ref{eq:coriolis}). 
%If the vehicles are only allowed to move at low speeds and have three planes of symmetry, the off-diagonal elements of the added matrix can be neglected. This approximation is often suitable for practical applications as the off-diagonal elements are typically much smaller than the diagonal elements.

The hydrodynamic damping matrix, $D(\nu_r)$, accounts for the forces due to various hydrodynamic effects, including potential damping, skin friction, wave drift damping, vortex shedding, and lifting forces. It is expressed as a sum of linear and quadratic terms, where the linear damping matrix $D_l$ is constant, and the quadratic damping matrix $D_n$ is proportional to the absolute value of the relative velocity $|\nu_r|$. The dominance of linear or nonlinear damping depends on the surge velocity of the vehicle.

In the case of noncoupled motion, where the motion in one axis does not affect the other axes, a diagonal damping structure can be assumed. This simplifies the hydrodynamic damping matrix, which can be expressed as:

\begin{align}
\begin{split}
    D(\nu_r) = &-\diag([X_u, Y_v, Z_w, K_p, M_q, N_r]^\transpose) \\
    &- \diag ([X_{|u|u}|u_r|, Y_{|v|v}|v_r|, Z_{|w|w}|w_r|, \\ 
    & \;\;\;\;\;\;\;\;\;\;\;\;\; K_{|p|p}|p_r|, M_{|q|q}|q_r|, N_{|r|r}|r_r|]^\transpose),
    \end{split}
    \label{eq:damping}
\end{align}
where $\diag(x)$ denotes a diagonal matrix where diagonal elements are composed of elements of vector $x$. \cc{The damping allows the vehicle to dissipate energy, ensuring stability in both roll and pitch motions.}

%\tsl{We need to say this using some surface vehicle knowledge.}
% \cc{
% \begin{remark}
%     We need some clarification about the stability of the uncontrolled model. The damping $D$ enables the robot to dissipate energy and converge to zero pitch and roll angle.
% \end{remark}}
\cc{For establishing a practical controller, we assume that the vehicle is neutrally buoyant, thereby neglecting the hydrostatic force $g(\eta)$. Additionally, we consider the external disturbances caused by current, wind and waves are either unknown or negligible, and treat them as noise or modeling errors.}

\begin{equation}
M\dot{\nu} + C(\nu)\nu + D(\nu) \nu = \tau,
\label{eq:dyn-simple}
\end{equation}
where $M = M_{RB}+M_{AM}$ and $C = C_{RB} + C_{AD}$.

It is important to note that while our controller uses this simplified model and does not explicitly account for the external forces, in the simulation, we will incorporate an external force to assess the ability of the controller to handle significant disturbances. 

\subsection{Dynamics on a Lie group}
This section briefly introduces the dynamics model on a Lie group $\mathrm{SE}(3)$ and the commonly used notation. For a more comprehensive understanding of Lie groups, refer to \cite{chirikjian2011stochastic,hall2015lie,sola2018micro,bullo2019geometric}. A group is a set, $\mathcal{G}$, with a composition operation for its elements that satisfies the axioms of closure, identity, inverse, and associativity. In a Lie group, the manifold is symmetric and looks the same at every point, making all tangent spaces isomorphic. The tangent space at the identity (since all groups have the identity element), $T_{e} \mathcal{G} = \mathfrak{g}$, is defined as the Lie algebra of $\mathcal{G}$. The exponential map, $\exp : \mathfrak{g} \rightarrow \mathcal{G}$, maps elements of the Lie algebra to elements of the Lie group. The inverse operation of the exponential map is the $\log$ map.

The Lie algebra is a vector space whose elements can be associated with vectors $\xi \in \mathbb{R}^n$, where $\dim \mathfrak{g} = n$. The conversion between $\mathfrak{g}$ and $\mathbb{R}^n$ is facilitated by the following isomorphism, commonly known as the hat and vee operators:

\begin{equation}
(\xi)^\wedge : \mathbb{R}^n \rightarrow \mathfrak{g}, \;\;
(\xi)^\vee : \mathfrak{g} \rightarrow \mathbb{R}^n.
\end{equation}

In matrix Lie groups, the $\exp$ map naturally arises by exactly integrating the group reconstruction equation,
\begin{equation}
\dot{X} = X \xi^\wedge .
\end{equation}
% where $\mathbf{v}$ is the velocity, $\xi^\wedge = (\mathbf{v}t)^\wedge = \mathbf{v}^\wedge t$. \mgj{I think it shouldn't be bold $v$. Time wasn't defined as $t$.}

The vehicle state in $\mathrm{SE}(3)$ can be represented by a rotation matrix $R \in \mathrm{SO}(3)$ and position $p \in \mathbb{R}^3$. The homogeneous representation of an $\mathrm{SE}(3)$ element is given by:
\begin{equation}
X = \begin{bmatrix}
R & p  \\
0 & 1 
\end{bmatrix}.
\end{equation}

We define the twist as the concatenation of the angular velocity $\omega$ and the linear velocity $v$ in the body frame, denoted as $\xi = [\nu_2^\transpose, \nu_1^\transpose]^\transpose =  [\omega^\transpose, v^\transpose]^\transpose \in \mathbb{R}^6$. The hat operator is then used to obtain the corresponding $T_{e} \mathrm{SE}(3) = \mathfrak{se}(3)$ element:

\begin{equation}
\xi^\wedge = \begin{bmatrix}
\omega^\wedge     & v  \\
0 & 0 
\end{bmatrix}.
\end{equation}

For an $X \in \mathrm{SE}(3)$, it can be shown that both $X^{-1}\dot{X}$ and $\dot{X}X^{-1}$ belong to $\mathfrak{se}(3)$. The former is the body velocity in the body-fixed frame, while the latter is the spatial velocity in the spatial frame. The relationship between these two velocities is given by the adjoint map, $\mathrm{Ad}_X : \mathfrak{g} \rightarrow \mathfrak {g}$, that enables change of frame for velocities defined in the Lie algebra via the following matrix similarity.
\begin{equation}
\mathrm{Ad}_X \xi = X \xi^\wedge X^{-1} . 
\end{equation}

% \mgj{Above should be $\mathrm{Ad}_X \xi = X \xi^\wedge X^{-1}$ for consistency. $\xi$ as defined as a vector previously.}

The adjoint map describes how elements of a Lie group or a Lie algebra act on other elements of a Lie algebra. The adjoint map is a linear transformation that maps an element $\xi\in\mathfrak{g}$ to $\mathrm{Ad}_X(\xi)$. 
% On the other hand, the adjoint of an element of the Lie algebra is an intrinsic property of a Lie algebra itself and does not depend on any specific Lie groups. 
For an element $\xi\in\mathfrak{g}$, its derivative at the identity denoted $\mathrm{ad}_\xi:\mathfrak{g}\rightarrow\mathfrak{g}$ and it maps an element from $\eta\in\mathfrak{g}$ to $\mathrm{ad}_\xi(\eta)$. The  (little) adjoint describes how the Lie bracket acts on an element of the Lie algebra. The Lie bracket is a bilinear operation on the Lie algebra that measures the failure of the product of two Lie group elements to commute (Lie derivative). 

% If we apply the adjoint map $\mathrm{Ad}_{\exp(\xi)}$ to an element $\eta\in\mathfrak{g}$, we get the adjoint of $\xi$ acting on $\eta$:

The adjoint map and adjoint in the Lie algebra in $\mathrm{SE}(3)$ can be represented by matrices as follows:

\begin{equation}
\mathrm{Ad}_X= \begin{bmatrix}
R   & 0  \\
p^\wedge R & R 
\end{bmatrix}, \;\; 
\mathrm{ad}_\xi = \begin{bmatrix}
\omega^\wedge & 0    \\
v^\wedge & \omega^\wedge    
\end{bmatrix}.
\end{equation}

Euler-Poincar\'e equations \cite{bloch1996euler} is aligned with the hydrodynamic equation (\ref{eq:dyn-simple}) as:

\begin{align} \label{eq:dyn-lie}
\begin{split}
M\dot{\xi} &= \mathrm{ad}_\xi ^\transpose M \xi + f \\
&=-C(\xi)\xi  - D(\xi) \xi + \tau,
\end{split}
\end{align}

\begin{equation}
\begin{bmatrix}
\dot{R}   & \dot{p}  \\
0 & 0
\end{bmatrix} =
\begin{bmatrix}
 R  & p  \\
0 & 1
\end{bmatrix}
\begin{bmatrix}
 \omega^\wedge  & v  \\
0 & 0
\end{bmatrix}.
\end{equation}
where $f \in \mathfrak{g}^*$~\footnote{Technically, quantities that depend on mass and inertial belong to the co-tangent space $\mathfrak{g}^*$~\cite{bloch2015nonholonomic}.} is the external force applied to the body fixed principal axes, including damping force and control force, i.e., $f:= - D(\xi) \xi + \tau$. \cc{The dynamics model, defined using Lie groups, is applied to a state that resides on a symmetric manifold, thereby ensuring its preservation in a globally identical form.}

Note that the variables $\nu$ and $\xi$ are equivalent, but they differ in the order of their elements. Consequently, the matrices $M$, $C$, and $D$ have different orderings.

\section{Geometric convex error-state MPC}
\label{Geometric convex error-state MPC}

\cc{In this section, we develop a convex error-state MPC by linearizing the error and motion dynamics. The use of error defined in Lie algebra enables the linearzed error dynamics to be independent from the orientation of the robot, thus rendering it suitable for long-term prediction. Additionally, we will demonstrate that the nonlinearity of the vehicle dynamics model can also be readily accommodated in convex form through linearization, resulting in a versatile and generalized convex MPC problem.}

We consider a desired trajectory in the Lie group $\mathcal{G}$ as a function of time $t$, denoted $X_{d,t} \in \mathcal{G}$. We define the left-invariant error~\cite{barrau2017invariant} $\Psi$ and its dynamics as:
\begin{equation}
\Psi = X_{d,t}^{-1} X_t \in \mathcal{G},
\end{equation}
\begin{equation}
\frac{d}{dt} \Psi = \frac{d}{dt}(X_{d,t}^{-1})X_t + X_{d,t}^{-1} \frac{d}{dt}X_t = \Psi_t \xi_t^\wedge - 
\xi_{d,t}^\wedge \Psi_t.
\end{equation}
where $\xi_t$ and $\xi_{d,t}$ are the velocity vectors corresponding to $X_t$ and $X_{d,t}$, respectively.

To compare velocities from different reference frames, we use the transport adjoint map $\mathrm{Ad}_{\Psi}$ and obtain the error dynamics as:
\begin{equation}
\dot{\Psi} = \Psi_t (\xi^\wedge - \Psi_t^{-1} \xi_{d,t}^\wedge \Psi_t) = \Psi (\xi_t - \mathrm{Ad}_{\Psi_t^{-1}} \xi_{d,t})^\wedge .
\end{equation}

Given the first-order approximation of the exponential map, we define the error in the Lie algebra corresponding to $\Psi_t$ as:
\begin{equation}
\Psi_t = \exp(\psi_t^\wedge) \approx I + \psi_t^\wedge .
\end{equation}

We then obtain the linearized error dynamics in the Lie algebra as:
\begin{equation}
\dot{\Psi}_t \approx \dot{\psi_t}^\wedge \approx (I+\psi_t^\wedge ) (\xi_t - \mathrm{Ad}_{I-\psi_t^\wedge} \xi_{d,t})^\wedge .
\end{equation}
Here, $\psi_t$ is the corresponding error in the Lie algebra for $\Psi_t$, and we use the property $\mathrm{Ad}_{\Psi} = \exp (\mathrm{ad}_{\psi})$. Given a first-order approximation, $\mathrm{Ad}_{I+\psi^\wedge} = I + \mathrm{ad}_{\psi}$.
Finally, we obtain the linearized velocity error in the Lie algebra as:
\begin{equation} \label{eq:linearized-vel-err}
\dot{\psi_t} = -\mathrm{ad}_{\xi_{d,t}}\psi_t + \xi_t - \xi_{d,t} .
\end{equation}

Since we now have a linear model for the error dynamics, we proceed with the linearization of the hydrodynamics described in (\ref{eq:dyn-lie}). The linearization is performed around the operating point $\bar{\xi}$:
\begin{align}
\begin{split}
M\dot{\xi} &\approx -C(\bar{\xi})\bar{\xi} -D(\bar{\xi})\bar{\xi} \\
& \;\;\;\;\; - \frac{\partial C(\xi)\xi}{\partial \xi}|_{\bar{\xi}}(\xi - \bar{\xi}) - \frac{\partial D(\xi)\xi}{\partial \xi}|_{\bar{\xi}}(\xi - \bar{\xi}) + \tau \\
&= (-C(\bar{\xi}) - D(\bar{\xi}) - \frac{\partial C(\xi) \bar{\xi}}{\partial \xi} - \frac{\partial D(\xi) \bar{\xi}}{\partial \xi}) \xi \\ 
&\;\;\;\;\; + (\frac{\partial C(\xi) \bar{\xi}}{\partial \xi} + \frac{\partial D(\xi) \bar{\xi}}{\partial \xi} )\bar{\xi} + \tau \\
&=H_t \xi + b_t .
\end{split}
\end{align}

% \mgj{Let's add an extra dissipation term to (24) evaluated at the desired twist $\xi_d$. $$M\dot{\xi} = H_t \xi + b_t + \alpha \cdot \mathrm{ad}^\transpose_{M\xi_d}(\mathrm{ad}^\transpose_{M\xi_d}(\xi_d)).$$
% $\alpha \in \mathbb{R}_{>0}$ is a tunable parameter for scaling.}

As shown in (\ref{eq:coriolis}) and  (\ref{eq:damping}), $C(\xi)$ and $D(\xi)$ are generally represented as first-order functions of $\xi$, so partial differentiation will yield a constant. In such cases,
\begin{align}
\begin{split}
&\frac{\partial C(\xi) \bar{\xi}}{\partial \xi} \\
&=  -\frac{\partial }{\partial \xi} 
   (\begin{bmatrix}
        (M_{21}v + M_{22}\omega)^\wedge & (M_{11}v + M_{12}\omega)^\wedge \\
        (M_{11}v + M_{12}\omega)^\wedge & 0 
    \end{bmatrix}
    \begin{bmatrix}
        \bar{\omega} \\
        \bar{v}
    \end{bmatrix}) \\
    &=\frac{\partial }{\partial \xi} ( \begin{bmatrix}
        \bar{\omega}^\wedge M_{22} + \bar{v}^\wedge M_{12} & \bar{\omega}^\wedge M_{21} +\bar{v}^\wedge M_{11} \\
        \bar{\omega}^\wedge  M_{12} & \bar{\omega}^\wedge M_{11}
    \end{bmatrix}
    \xi) \\
    &=\begin{bmatrix}
        \bar{\omega}^\wedge M_{22} + \bar{v}^\wedge M_{12} & \bar{\omega}^\wedge M_{21} +\bar{v}^\wedge M_{11} \\
        \bar{\omega}^\wedge  M_{12} & \bar{\omega}^\wedge M_{11}
    \end{bmatrix},
\end{split}
\label{eq:coriolis-diff}
\end{align}

\begin{equation}
\begin{split}
\frac{\partial D(\xi) \bar{\xi}}{\partial \xi} =- \diag ([ X_{|u|u}|\bar{u}|, Y_{|v|v}|\bar{v}|, Z_{|w|w}|\bar{w}|, \\
K_{|p|p}|\bar{p}|, M_{|q|q}|\bar{q}|, N_{|r|r}|\bar{r}|]^\transpose )  .
\end{split}
\end{equation}

% \mgj{The variables are different? We don't have roll pitch yaw.}

% \jw{Which variables you mean? We have roll pitch yaw for both C and D, ($\bar{\omega}$) and (p, q, r)}

We define the system states as $x_t = [\psi_t, \xi_t]^\transpose$. Then, the linearized dynamics become:

\begin{equation}
    \dot{x_t} = A_t x_t + B_t \tau + h_t,
\end{equation}
where
\begin{equation}
    A_t = \begin{bmatrix}
        -\mathrm{ad}_{\xi_{d,t}} & I \\
        0 & H_t
    \end{bmatrix},
    B_t = \begin{bmatrix}
        0 \\
        M^{-1}
    \end{bmatrix},
    h_t = \begin{bmatrix}
        -\xi_{d,t} \\
        b_t
    \end{bmatrix}.
\end{equation}

We set the operating point $\bar{\xi}$ to be the reference trajectory $\xi_{d,t}$ and set the cost function to regulate the tracking error $\psi_t$ and its derivative $\dot{\psi}_t$, rather than the difference between $\xi_{d,t}$ and $\xi_t$. With the tracking error defined as $y_t = [\psi_t^\transpose, \dot{\psi}_t^\transpose]^\transpose$, 

\begin{equation}
    y_t = G_t x_t - d_t
\end{equation}
\begin{equation}
    G_t =\begin{bmatrix}
        I & 0 \\ -\mathrm{ad}_{\xi_{d,t}} & 0
    \end{bmatrix},
    d_t = \begin{bmatrix}
        0 \\ \xi_{d,t}
    \end{bmatrix},
\end{equation}
the cost function is formulated as follows. 
% \mgj{check the notation $u$ vs. $\tau$ and indexing for time ($y_t,u_t$).}

\cc{
\begin{equation}
    J = y_{t_H}^\transpose P y_{t_H} + \int_{t=0}^{t_H} (y_t^\transpose Q y_t + \tau_t^\transpose R \tau_t) dt,
\end{equation}
where $t_H$ is the length of the prediction horizon time in MPC}, and $P$, $Q$, and $R$ are semi-positive definite cost matrices. \cc{After discretizing the system given the time step $\{t_k\}_{k=1}^{N}$}, we can construct a QP problem that can be solved efficiently using a QP solver such as OSQP \cite{stellato2020osqp}.

\begin{problem}
    (Proposed MPC) Find $\tau_k \in \mathfrak{g}^*$ such that 
    \begin{align*}
        \min_{\tau_k}  & \;\;\;\; y_N^\transpose P y_N + \sum_{k=1}^{N-1} y_k^\transpose Q y_k  + \tau_k^\transpose R \tau_k \\
        s.t.  \;\; & x_{k+1} = A_k x_k + B_k \tau_k + h_k \\
         & x(0) = x_0, \;\; \tau_k \in \mathcal{T}_k, \;\;  k = 0, 1, ..., N-1 .
    \end{align*}
\end{problem}
where $\mathfrak{g}^*$ is the cotangent space, \cc{$\mathcal{T}$ is the feasible control force set, $A_k = I + A_{t_k} \Delta t$, $B_k = B_{t_k} \Delta t$}, and $h_k = h_t \Delta t$.

\section{Numerical Simulations}
\label{sec:Numerical Simulations}

We evaluate the performance of our controller by applying it to a marine vehicle simulator. To thoroughly evaluate its robustness, we compare our controller with NMPC methods while considering the presence of external disturbances.

\subsection{Surface vehicle dynamics model}
We validate the control performance of the proposed algorithm using Marine Systems Simulator \cite{fossen2004}. The simulator has hydrodynamic models of various types of real-world vehicles, including the autonomous surface vehicle Otter. The Otter is a 2 m catamaran equipped with two propellers on the starboard and port sides, allowing it to achieve a maximum speed of 5.5 knots.

\cc{The Otter model is suitable for demonstration because it is defined in 6-DOF and accounts for hydrostatic forces.}  In the hydrodynamics model of the Otter, the Coriolis matrices $C_{RB}$ and $C_{AM}$ are set differently. Specifically, the elements in $C_{AM}$ related to the yaw angle and horizontal velocity are neglected. To handle this, we separately calculate the partial differentiations (\ref{eq:coriolis-diff}) of the Coriolis matrices. The damping matrix of the Otter model is set to be diagonal with an additional nonlinear term only for the yaw motion. 

The Otter is an under-actuated system that requires controlling its position and orientation using only two control inputs, the rotational speeds of the two motors. The propulsion force generated by each motor is modeled as proportional to the square of the rotational speed, with the coefficient changing for reverse motion. Although this relationship is nonlinear, it is bijective, allowing us to use the propulsion forces as a control input. 
\begin{equation}
    \tau = Tu = 
    \begin{bmatrix}
    0 & 0 & l & 1 & 0 & 0 \\
    0 & 0 & -l & 1 & 0 & 0 
    \end{bmatrix}^\transpose
    \begin{bmatrix}
        u_1 \\ u_2
    \end{bmatrix},
\end{equation}
where $u_1$ and $u_2$ represent the port and starboard thrust forces, respectively, and $l$ denotes the distance from the thrusters to the center of gravity along the y-axis.

\subsection{Nonlinear MPC}
Due to the intricate hydrodynamics model of the vehicle, nonlinear optimization is required for a controller, which can be time-consuming. Contrarily, the proposed geometric MPC provides computational efficiency by formulating the problem as a convex QP problem. However, the employment of linearization in the proposed algorithm has the potential to undermine control performance. Therefore, we consider NMPC as the baseline and regard its tracking performance as the benchmark for comparison with our method.

The NMPC algorithm was implemented using CasADi \cite{Andersson2019} \cc{with MPCTools \cite{risbeck2016}}, which is an open-source software tool for nonlinear optimization and algorithmic differentiation.  We distinguish between two forms of NMPC: the original form, referred to as \textit{NMPC}, which considers hydrostatic forces; and \textit{NMPC-simple}, which utilizes a simplified model (\ref{eq:dyn-simple}) equivalent to the model employed in the proposed method. We assume that no information pertaining external disturbances, such as ocean current speed, is available. Therefore, we use the model with $\nu_c=0$ to optimize the control inputs. 

Let the error variables be denoted as $z_t = ([\eta_{d,t}^\transpose, \xi_{d,t}^\transpose] - [\eta_t^\transpose, \xi_t^\transpose])^\transpose$, then \textit{NMPC} is defined as follows:

\begin{problem}
    (NMPC) Find $u_k \in \mathbb{R}$ such that 
    \begin{align*}
        \min_{u_k}  & \;\;\;\; z_k^\transpose P z_k + \sum_{k=1}^{N-1} z_k^\transpose Q z_k  + u_k^\transpose R u_k \\
        s.t.  \;\; & \xi_{k+1} = M^{-1} (-C(\xi_k)\xi_k - D(\xi_k)\xi_k - g(\eta_k) + Tu_k) \\
         & \;\;\;\; \eta_{k+1} = J_{\Theta}(\eta_k) \xi_k \\ 
         & \eta(0) = \eta_0, \;\; \xi(0) = \xi_0, \;\; u_k \in \mathcal{U}_k, \;\; k = 0, 1, ..., N-1 \;, 
    \end{align*}
\end{problem}
where $\mathcal{U}$ is the feasible control input set.

\textit{NMPC-simple} is an equivalent optimization problem, but it does not take into account the hydrostatic term $g(\eta)$.

% \begin{problem}
%     (NMPC-simple) Find $u_k \in \mathbb{R}$ such that 
%     \begin{align*}
%         \min_{u_k}  & \;\;\;\; z_k^\transpose P z_k + \sum_{k=1}^{N-1} z_k^\transpose Q z_k  + u_k^\transpose R u_k  \\
%         s.t.  \;\; & \xi_{k+1} = M^{-1} (-C(\xi_k)\xi_k - D(\xi_k)\xi_k  + Tu_k) \\
%         & \;\;\;\; \eta_{k+1} = J_{\Theta}(\eta_k) \xi_k \\ 
%          & \eta(0) = \eta_0, \;\; \xi(0) = \xi_0, \;\; k = 0, 1, ..., N-1
%     \end{align*}
% \end{problem}

\subsection{Simulation setup}
The control frequency of the MPC was set to 20 Hz, while the simulation frequency was set to 80 Hz to capture the nonlinear dynamics of the vehicle accurately. The MPC horizon length is chosen as 100 steps (i.e., $N = 100$), which corresponds to predicting 5 second future trajectories. \cc{The prediction horizon was determined to be the minimum length required to prevent control divergence.} Each simulation episode has a duration of 60 seconds. The MPC cost weights $P, Q, R$ are adjusted to facilitate smooth tracking of the desired trajectory with minimal steady-state error. We empirically found that while position error plays a significant role in achieving accurate tracking, considering angle and velocity errors help prevent divergence and fluctuations.

To assess the control performance of the vehicle, we evaluated the tracking accuracy during zigzag and constant turning maneuvers. The initial position and orientation of the vehicle are randomly placed within a 5 m radius. The desired trajectory has a surge speed of 0.5 m/s. For turning motion, a desired yaw velocity is 0.1 rad/s, and for zigzag motion, a desired yaw velocity is designed using a sinusoidal function, $0.1 \sin(t/5)$ rad/s.

All simulations were conducted with 10 random Monte Carlo tests and were performed using a Geforce GTX 1050 Ti, Intel(R) Core i7-8850H CPU @ 2.60 GHz $\times$ 12, a memory of 64 GB, and Ubuntu 22.04.

\subsection{Trajectory tracking in a current-free environment}
\begin{figure}[t]
     \centering
     \subfloat[]{
     \includegraphics[scale= 0.44]{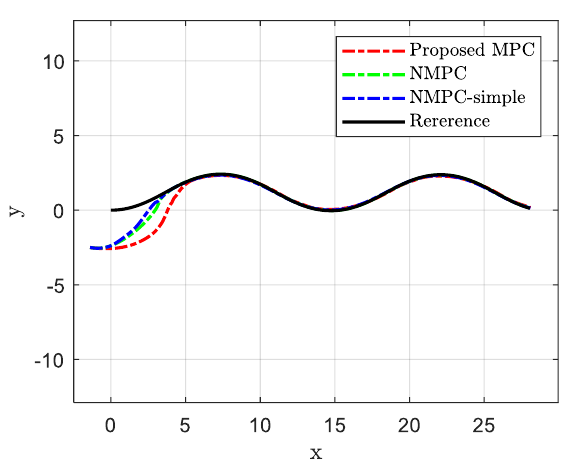}
     }
     % \\
     % \hspace{-5pt}
     \subfloat[]{
     \includegraphics[scale=0.44]{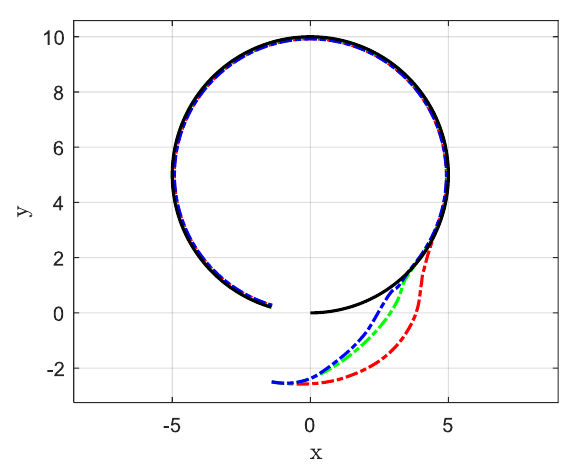}
     }
\caption{Simulation results in a current-free environment for Otter USV trajectory tracking in (a) zigzag motion and (b) turning motion. The reference path is represented by the solid black line, which includes the desired orientation and velocity over time. The controlled trajectories according to each method are represented by the dashed line. The proposed controller exhibits less sharp turning compared to NMPC controllers where rapid turning motion requires reducing surge speed. However, after the tracking error converges, the vehicle accurately follow the reference path.}
     \label{fig:traj}
% \vspace{-13pt}
\end{figure}

In a current-free environment, the known dynamics model used in MPC is equivalent to the simulation model except for the difference in sampling time. Figure \ref{fig:traj} illustrates the results of trajectory tracking control, aligned with the desired zigzag and turning trajectory. Direct optimization using precise nonlinear dynamics, such as \textit{NMPC} and \textit{NMPC-simple}, yields accurate optimization and rapid convergence. The proposed algorithm, which employs first-order approximations in the dynamics and error dynamics models, follows a roundabout path. Once convergence is achieved, the algorithm successfully tracks the predetermined reference path, similar to the NMPC methods.

%% figure

\begin{figure}[t]
\begin{centering}
\subfloat[]{
\includegraphics[scale=0.43]{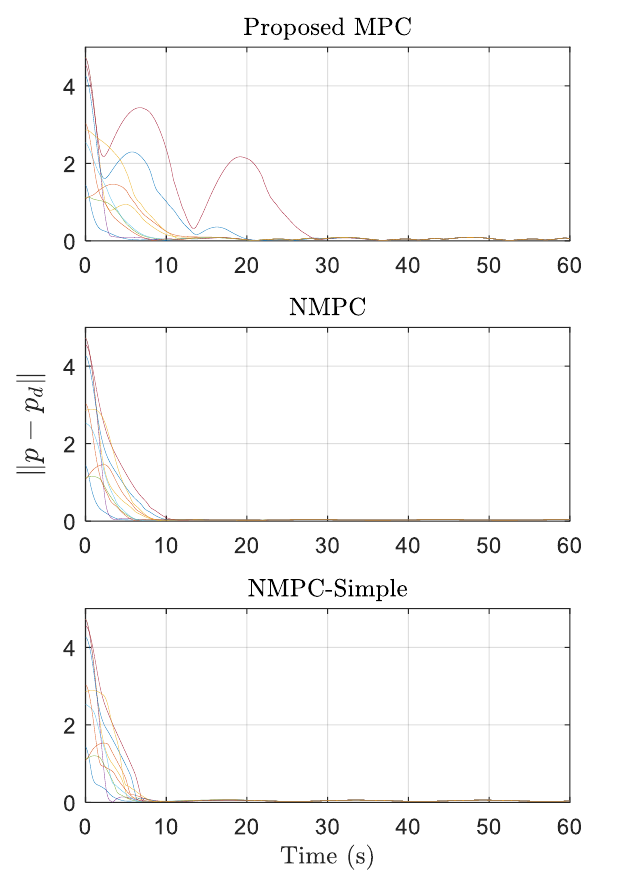}
}
\subfloat[]{
\includegraphics[scale=0.43]{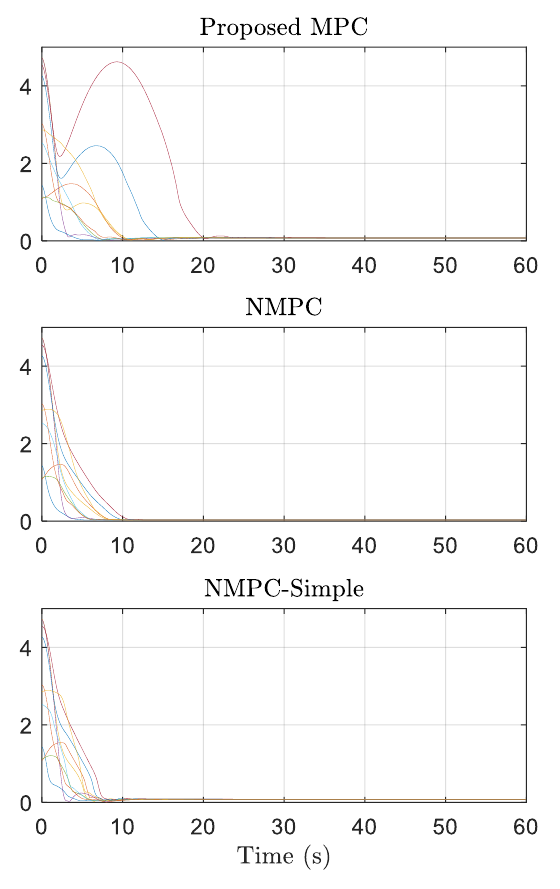}
}
\par\end{centering}
\caption{\label{fig:error_nocurrent} Tracking errors of the controllers evaluated in a current-free environment using 10 randomly sampled initial states for (a) zigzag motion and (b) turning motion. Our controller may exhibit overshoot; however, the performance difference is negligible once it converges.}
% \vspace{-13pt}
\end{figure}

Fig.~\ref{fig:error_nocurrent} shows the positional errors over time for 10 episodes in each maneuver. Both \textit{NMPC} and \textit{NMPC-simple} achieve fast convergence, and their performance are nearly indistinguishable. The proposed algorithm performs similarly to NMPC methods in cases with small initial errors. However, for larger initial errors, the algorithm exhibits some overshoot, resulting in longer stabilization times. Nevertheless, the algorithm converges within 30 seconds in all cases and successfully performs trajectory tracking. The \textit{proposed MPC} and \textit{NMPC-simple} exhibits a small position error bias since they do not use perfect dynamics, although the value is very marginal, less than 0.1 m.

%% figure

\subsection{Trajectory tracking in a current-carrying environment}
\begin{figure}[t]
     \centering
     \subfloat[]{
     \includegraphics[scale=0.44]{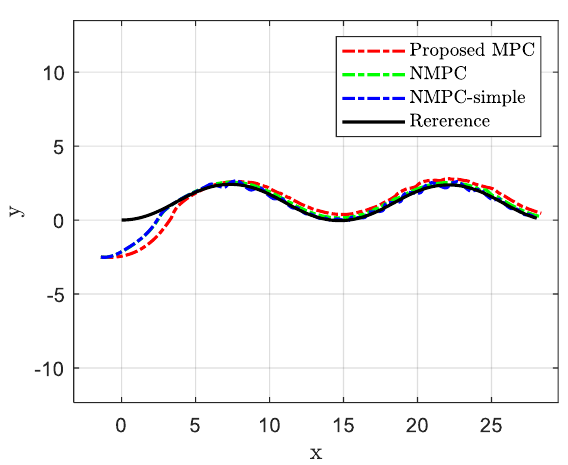}
     }
     % \\
     % \vspace{-10pt}
     \subfloat[]{
     \includegraphics[scale=0.44]{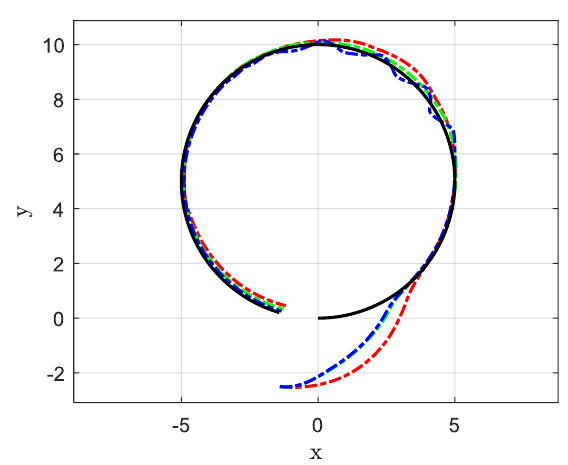}
     }
     \caption{Simulation results under an ocean current speed of 0.5 m/s for (a) zigzag motion and (b) turning motion. \textit{NMPC} shows a smooth trajectory with low steady-state error. While both \textit{proposed MPC} and \textit{NMPC-simple} employ the same approximated dynamics model, \textit{propsed MPC} generates a smoother trajectory during turning maneuvers.}
     \label{fig:traj_current}

% \vspace{-13pt}
\end{figure}

\begin{figure}[ht]
\begin{centering}
\subfloat[]{
\includegraphics[scale=0.43]{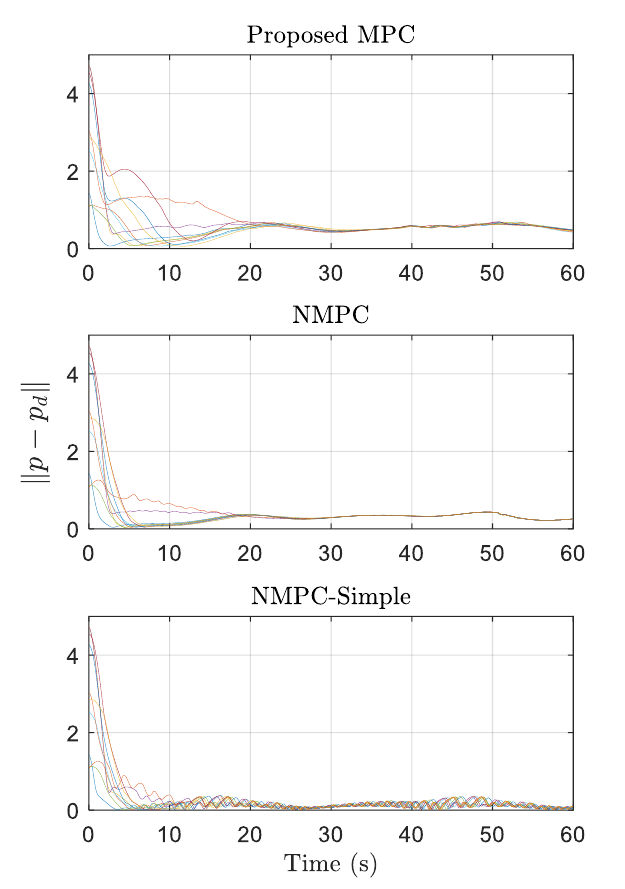}
}
\subfloat[]{
\includegraphics[scale=0.43]{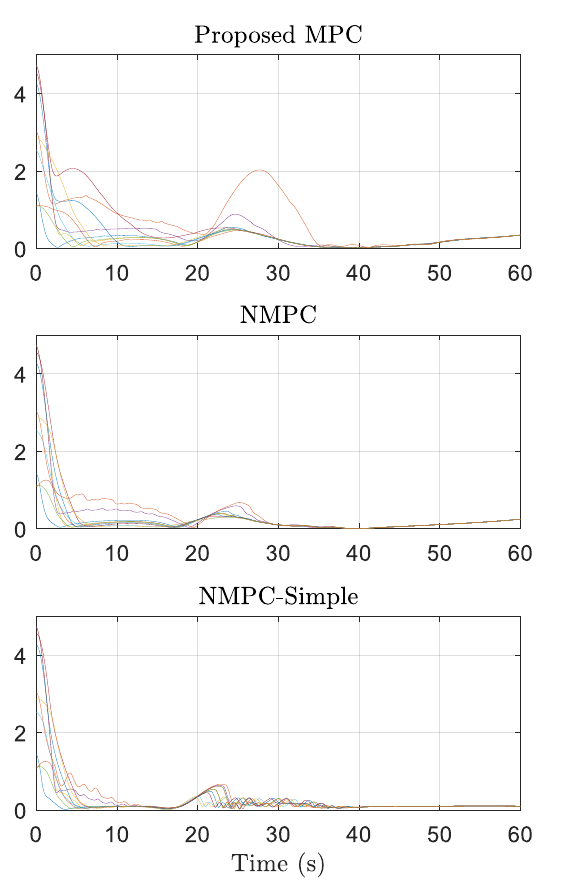}
}
\par\end{centering}
\caption{\label{fig:error_current} Tracking errors of the controllers under an ocean current speed of 0.5 m/s for (a) zigzag motion and (b) turning motion. Although the external disturbance is not accounted for in MPC algorithms, MPC can robustly respond to the disturbance and successfully track the desired trajectory, albeit with some steady-state errors.}
% \vspace{-13pt}
\end{figure}

To validate the applicability of the proposed algorithm in the presence of an external disturbance, experiments were conducted in an environment with the ocean current. The current is assumed to have a constant direction with speeds ranging from 0 to 0.5 m/s in 0.1 m/s increments. Despite the controller's lack of knowledge about the ocean current information, the iterative feedback of MPC enables robust system control in response to the disturbance. Fig.~\ref{fig:traj_current} and \ref{fig:error_current} depict the trajectories generated by each controller and tracking errors of 10 episodes, respectively, at an ocean current speed of 0.5 m/s. 

As observed in previous control results without the ocean current, NMPC methods exhibit aggressive turning compared to the \textit{Proposed MPC}, enabling them to converge to the desired path quickly. However, this aggressive turning behavior can lead to fluctuations in the presence of modeling gaps. \textit{NMPC-simple}, which utilizes a less accurate dynamics model, deviates more in terms of the vehicle's position and orientation from the desired path. When the orientation error becomes significant, it makes aggressive adjustments to the vehicle's position, resulting in a fluctuated trajectory. In contrast, although \textit{Proposed MPC} also employs an equivalent approximated model, it generates a smoother trajectory by utilizing linearized dynamics, which is desirable in practical applications.

%% figure

%% figure

\begin{figure}[ht]
\begin{centering}
\includegraphics[scale=0.73]{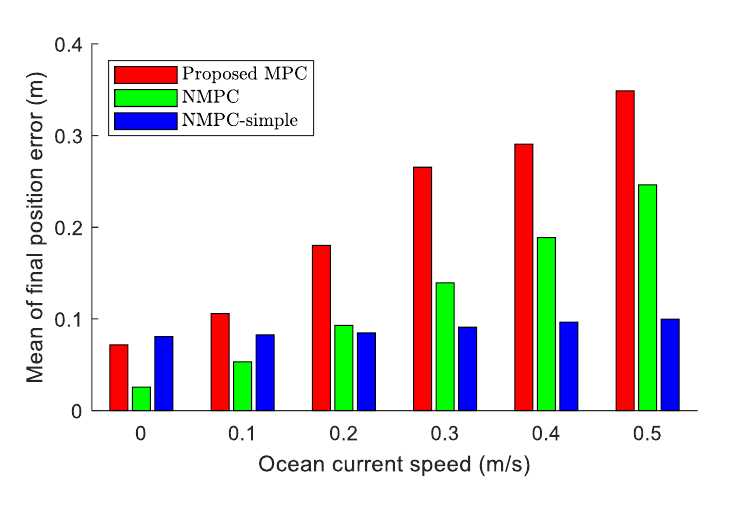}
\par\end{centering}
\caption{\label{fig:current_final_error} Average final position error in the turning motion when a constant ocean current occurs. As the speed of the ocean current increases, the final position error increases. The final error varies depending on the angle of encounter with the current; thus, these results are made when the final error is maximized. \textit{Proposed MPC} demonstrates the ability to control within an appropriate range of errors like \textit{NMPC}, even with relatively high ocean current speeds. 
%\mgj{I think we could add a boxplot to show the full distribution of the results. The mean is biased and is only part of the story. One empirical CDF plot to summarize the results would be interesting too. We have space and there are plenty of white spaces to remove from figures.}
% \mgj{After thinking more about the results, one of our claims is that we can increase the planning horizon, which is understood to be important. But we do know to show its effect. Another figure like this one to show error vs. planning horizon, 1,3, ..., max, would add lot.}
}
% \vspace{-13pt}
\end{figure}

Figure \ref{fig:current_final_error} illustrates the average final position error in the turning maneuver at each ocean current speed. The position error at the final position increases with the current speed except for \textit{NMPC-simple}. As \textit{NMPC-simple} produces fluctuated vehicle trajectory, the final position error is not significantly affected by the ocean current. \textit{Propsed MPC} exhibits slightly higher errors compared to \textit{NMPC}. Nevertheless, given the low maneuverability of the marine vehicle, the maximum position error not exceeding 0.4 m even at high ocean currents of 0.5 m/s indicates that the proposed algorithm demonstrates sufficiently high control performance.

%% figure

While the tracking performance of the proposed MPC is comparable to that of NMPC, the proposed algorithm provides a significant computational efficiency advantage. Table~\ref{tab:comptime} presents the average time taken for each optimization. 
The optimization time encompasses all the processes necessary for generating control inputs, including formulating a problem, constructing a solver, and performing optimization. In the case of the \textit{Proposed MPC} with OSQP, a solver needs to be built in each iteration, whereas NMPC methods utilizing CasADi build the problem once and update problem parameters to minimize the overall computation time. In NMPC methods, a warm start that utilizes the solution from the previous iteration as an initial guess is employed to expedite the optimization process.
The proposed algorithm can run at 20 Hz, while NMPC methods can run at 1 to 2 Hz. \textit{Proposed MPC} requires approximately 10 times less computation time compared to NMPC methods, even when using simplified dynamics. This indicates that using NMPC is difficult for real-time control in such long-horizon prediction problems. 

%While the proposed algorithm took around 60 seconds for simulation with a 20 Hz control frequency, NMPC methods consumed around 10 times as much computation time as the proposed algorithm, even when using simplified dynamics. This indicates that using NMPC is difficult for real-time control in such long-horizon prediction problems. 

% \tsl{The runtime for the casadi may need more clarification. Did you formulate the problem instances every time in the control loop or you initialize a single instance of the controller and only update the parameters? The former may involve redundant time in symbolic computation. Did you use warm start to make the NMPC faster? }

\begin{table}[h]
\caption{\label{tab:comptime} Average computation time (ms) for single optimization.}
\centering
\begin{tabular}{llll}
\toprule
Current        & Proposed MPC & NMPC  & NMPC-simple \\ \midrule
0 m/s & \textbf{49 ($\pm2$)}         &  764 ($\pm13$) & 478 ($\pm12$)     \\
0.5 m/s & \textbf{50 ($\pm1$)}         &  953 ($\pm125$) & 460 ($\pm11$)    \\
 \bottomrule
\end{tabular}
% \vspace{-10pt}
\end{table}

% \begin{table}[h]
% \caption{\label{tab:comptime} Average computation time (sec) for single optimization.}
% % \renewcommand{\arraystretch}{1.1} 
% \centering
% \begin{tabular}{llll}
% \toprule
% Current        & Proposed MPC & NMPC  & NMPC-simple \\ \midrule
% 0 m/s & \textbf{48.3 ($\pm2$)}         & 917 ($\pm16$) & 574 ($\pm14$)    \\
% 0.5 m/s & \textbf{60 ($\pm2$)}         & 1143 ($\pm150$) & 552 ($\pm14$)    \\
%  \bottomrule
% \end{tabular}
% \vspace{-10pt}
% \end{table}

\subsection{Discussions}
The proposed algorithm leverages the geometric properties of the Lie group to define error dynamics and construct a convex error-state MPC for rapid control optimization. Although it exhibits slightly inferior trajectory tracking performance compared to NMPC \cc{due to linearization}, it offers significant reductions in computation time, rendering it suitable for real-time control in scenarios requiring long prediction horizons or involving a small marine vehicle. We may assume knowledge of the initial position and orientation of the vehicle, and the path planning algorithm can generate a reference path based on this; thereby, slow convergence at large initial errors can be mitigated in practical applications. 

QP-based optimal control problems offer advantages over nonlinear optimization for control performance, \cc{feasibility}, and stability analysis. \cc{As the proposed MPC only considers the input constraints, it does not suffer from feasibility issues compared to methods with state constraints.} \cc{To verify the stability of the proposed algorithm, the quadratic cost function in exponential coordinates \cite{teng2022lie} can be used as a candidate Lyapunov function. As we linearize the dynamics around the reference trajectory, the sum-of-square methods in \cite{tedrake2010lqr} can be applied to certify the region of attraction considering the control input constraints. Since the error-state MPC are derived in exponential coordinates, special treatment is required to incorporate \cite{teng2022lie} and \cite{tedrake2010lqr} for verification, which is an interesting future direction. }

%\tsl{Consider modifying the first sentence due to the added paragraph.}
\cc{Promising future work includes exploring stability and robustness where disturbances are incorporated into the model,} as demonstrated in tube-MPC \cite{zheng2020robust}, could be a valuable research direction. Our MPC assumes a control-affine system based on the forced Euler-Poincar\'e equation, where control inputs act linearly on the vehicle state. However, since a majority of large ships rely on a rudder system, which introduces nonlinearity between control inputs and actuated force, our method cannot be directly applied. Although linearizing the control force as with the dynamics model may be feasible, further investigation is required to ensure the effectiveness of our approach.

\section{Conclusion}
\label{sec:conclusion}
We propose a Lie algebraic convex error-state MPC algorithm, leveraging the intrinsic geometric properties of $\mathrm{SE}(3)$. We define the hydrodynamics of a marine vehicle on the Lie group and construct a convex error-state MPC by lifting the linearized problem to the Lie algebra. By using a globally valid model leveraged by the symmetry of tangent spaces in Lie groups, linearization of dynamics and error dynamics does not produce significant modeling gaps and can be optimized efficiently using a QP solver. Our results demonstrate that the proposed MPC has comparable tracking performance to nonlinear MPC but significantly reduces computation time, enabling real-time implementation with longer planner horizons. The proposed MPC is also robust to external disturbances such as ocean currents and generates smoother trajectories than NMPC using the same simplified hydrodynamics model. The proposed MPC will gain large benefits in real-time control applications.

{\footnotesize 
\balance
\bibliographystyle{IEEEtran}
\bibliography{strings-abrv,ieee-abrv,references}
}

\end{document}